\documentclass[11pt,a4paper]{article}
\usepackage[hyperref]{naaclhlt2018}
\usepackage{times}
\usepackage{latexsym}

\usepackage{url}

\aclfinalcopy 

\usepackage{amsmath,amssymb}
\DeclareMathOperator{\E}{\mathbb{E}}
\usepackage{graphicx}
\usepackage[labelsep=period]{caption}

\title{Generating Continuous Representations of Medical Texts}

\author{Graham Spinks \\
  Department of Computer Science \\
  KU Leuven, Belgium \\  
  {\tt graham.spinks@cs.kuleuven.be} \\\And
  Marie-Francine Moens \\
  Department of Computer Science \\
  KU Leuven, Belgium \\ 
  {\tt sien.moens@cs.kuleuven.be} \\}

\date{}

\begin{document}

\maketitle
\begin{abstract}
	
	We present an architecture that generates medical texts while learning an informative, continuous representation with discriminative features. During training the input to the system is a dataset of captions for medical X-Rays. The acquired continuous representations are of particular interest for use in many machine learning techniques where the discrete and high-dimensional nature of textual input is an obstacle. We use an Adversarially Regularized Autoencoder to create realistic text in both an unconditional and conditional setting. We show that this technique is applicable to medical texts which often contain syntactic and domain-specific shorthands. A quantitative evaluation shows that we achieve a lower model perplexity than a traditional LSTM generator.

\end{abstract}

\section{Introduction}

The main focus of this paper is the generation of realistic samples with a similar quality to those in a training set of medical texts. At the same time, an informative, continuous representation is created from the textual input. 

Obtaining a good representation for medical texts may prove vital to building more sophisticated generative, discriminative or semantic models for the field. One of the obstacles is the discrete nature of text that makes it difficult to employ in many machine learning algorithms.  
This is the case for Generative Adversarial Networks (GANs), which are not adequate to generate text as it is difficult to backpropagate the error to discrete symbols \cite{goodfellow2016nips}.

The ability of GANs to learn the underlying distribution, rather than repeating examples in the training data, has led to the successful generation of intricate high-resolution samples in computer vision \cite{zhang2017stackgan}. Conditional GANs in particular, where the class or label is passed to both generator and discriminator, implicitly learn relevant ancillary information which leads to more detailed outputs \cite{gauthier2014conditional,mirza2014conditional}. If we had a better understanding of how to train GANs with discrete data, some of those developments might be directly applicable to detailed text generation applications---such as image caption generation, machine translation, simplification of text, and text summarization---especially when dealing with noisy texts. 

Another impediment is the nature of clinical data, which is often unstructured and not well-formed, yet commonly has a high and important information density. Textual reports often don't follow regular syntax rules and contain very specific medical terminology. Moreover, the amount of training data is often limited and each physician has a personal writing style. Simply reusing pre-trained continuous representations, such as vector-based word embeddings \cite{turian2010word}, is therefore not always feasible for medical datasets.

The approach to text generation has mainly been dominated by Long Short-Term Memory networks (LSTMs). While LSTMs are successful in creating realistic samples, no actionable smooth representation is created of the text and thus there are limited possibilities to manipulate or employ the representations in additional applications that require continuous inputs. While the creation of continuous representations of text usually involves an autoencoder, the results mostly lack enough semantic information to be particularly useful in an alternate task. 

\citet{kim2017adversarially} have shown how to achieve text generation with a continuous representation by implementing an Adversarially Regularized Autoencoder (ARAE). They combine the training of a rich discrete space autoencoder with recurrent neural networks (RNNs) and the training of more simple, fully connected networks to generate samples in the continuous space. With adversarial (GAN) training, both the distribution of the generated as well as the encoded samples are encouraged to converge. The outcome is that a smooth representation is learned as well as a generator that can build realistic samples in the continuous space.

In this paper, we explore this methodology in the context of medical texts, more specifically captions for chest X-Rays. Analogous to conditional GANs, we also extend the network of \citet{kim2017adversarially} by generating samples conditioned on categorical, medical labels (for example 'healthy'). We refer to this method as conditional ARAE. In a quantitative evaluation, the perplexity of the conditional ARAE outperforms both the unconditional ARAE as well as a traditional LSTM.

\begin{figure}
	\centering
	\includegraphics[width=0.8\linewidth]{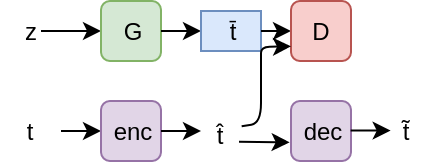}
	\caption[Overview of the architecture]{Overview of the ARAE architecture. The encoder $enc$ creates a new continuous representation $\hat{t}$ from the input text $t$. The decoder $dec$ tries to reconstruct the original text. Conjointly a generator $G$ and discriminator $D$ are trained in an adversarial setup. $z$ is a random normal noise vector. }
	\label{fig:architecture}
\end{figure}

\section{Methodology}
\begin{figure*}
	\centering
	\includegraphics[width=0.73\linewidth]{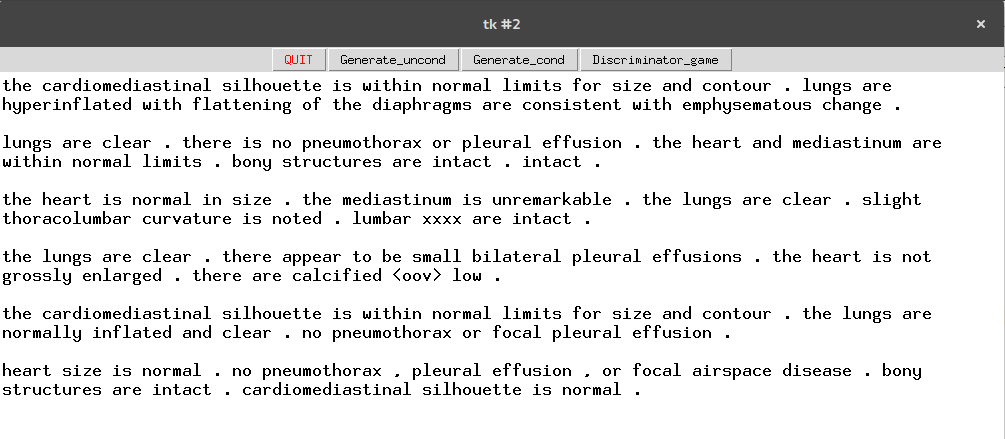}
	\caption[Demo screenshot]{A screenshot of the demo. Random unconditional samples are generated from a noise vector $z$. }
	\label{fig:demo_uncond}
\end{figure*}

For this demo we use the chest X-Ray dataset from the Indiana University hospital network \cite{demner2015preparing}. The dataset consists of 7470 X-Ray images for 3851 patients with corresponding textual findings. In this paper, the images are not used. We retain a maximum of 30 words per caption and pad shorter sentences. All words are transformed to lowercase and those with a frequency below 5 are removed and replaced by an out-of-vocabulary marker. The dataset also contains Medical Subject Headings (MeSH) which are essentially labels that indicate the main diagnoses for each patient report.

\subsection{GAN}

In a GAN, the loss is defined as a two-player min-max game between the generator ($G$) and the discriminator ($D$). The discriminator gradually improves its capacity to distinguish a sample from the real distribution from one of the generated distribution. The generator is trained on its ability to fool the discriminator into classifying its output as real data. The loss function can be described as:
\begin{multline}\label{eqn:eq1}
 \min\limits_{G} \max\limits_{D} \mathcal{L}_{GAN}(D,G) = \E_{x\sim p_{d}}[log(D(x))] + \\ \E_{z\sim p_{z}}[log(1 - D(G(z)))]  
\end{multline}
where $x$ represents a sample from the real data distribution $p_{d}$ and $z$ is a noise vector sampled from the distribution $p_{z}$, typically a random normal distribution, which serves as the input for $G$.  In a traditional GAN, the discriminator and generator are trained alternately in the hope that their performance improves each iteration. The setup of a traditional GAN  is essentially illustrated by the top row in figure \ref{fig:architecture}, where $\hat{t}$ would be a sample $x$ from the real distribution and $z$ is the noise vector.

To ensure convergence while training a GAN, we use an improved loss function, the Earth-Mover distance (EM)  \cite{arjovsky2017wasserstein}. The formulation of a GAN in equation \ref{eqn:eq1} minimizes the Kullback-Leibler divergence (KL-divergence) between the real and generated distributions. This formulation leads to infinite losses when the distribution of the generator maps to zero probability in the support of the real distribution. The Earth-Mover distance defines a more sensible distance measure for the convergence of two distributions $x$ and $y$ as specified in equation \ref{eqn:emdist}. It can be interpreted as a measure of the minimal amount of effort that is needed to move the mass of one distribution to match that of another. 
\begin{equation}\label{eqn:emdist}
W(P_r,P_g) = \inf_{\gamma \in \Pi(P_r ,P_g)} \mathbb{E}_{(x, y) \sim \gamma}\big[\:\|x - y\|\:\big]
\end{equation}
where $\Pi(P_r, P_g)$ is the set of all joint distributions $\gamma$ with marginal distributions $P_r$ and $P_g$. Computing the EM distance exactly is not tractable but \citet{arjovsky2017wasserstein} show that the Wasserstein GAN (WGAN) formulation (equation \ref{eqn:wgan}) leads to a theoretically sound and practical optimization of the EM distance if the parameters of D are restricted to a set of 1-Lipschitz functions, which in practice requires the weights to be constrained to an interval [-c,c]. In our experiments, c was set to 1 rather than a smaller value as we observe faster convergence during training. 
\begin{multline}\label{eqn:wgan}
\min\limits_{G} \max\limits_{D} \mathcal{L}_{W}(D,G) = \E_{x\sim p_{d}}[D(x)] - \\ \E_{z\sim p_{z}}[D(G(z))]  
\end{multline}

\subsection{ARAE} \label{captions}
In the traditional setup, GANs need to be fully differentiable to function. The error cannot adequately backpropagate when a series of discrete input variables are used. To alleviate this problem, an Adversarially Regularized Autoencoder (ARAE) is used \cite{kim2017adversarially}. The input text is mapped to a continuous representation $\hat{t}$ with a discrete, word-based autoencoder. The encoder $enc$ maps a text $t$ onto $\hat{t}$ while the decoder implements the conditional probability distribution $p(t|\hat{t})$. Both encoder and decoder are single-layer LSTMs. The loss is computed as the cross-entropy over the word reconstruction (equation \ref{eqn:loss_ce}).
\begin{equation}\label{eqn:loss_ce}
	\mathcal{L}_{rec}(enc,dec) = -log( \mathit{p}(t)|enc(t) )) 
\end{equation}
The discriminator of the ARAE now tries to determine which samples derive from the real distribution of encoded texts, $\hat{t}$, and which are generated from a noise distribution, $z$. While $D$ improves, the encoder enhances the representation $\hat{t}$ to contain more discriminative information.  To avoid divergence between encoder and decoder, only a portion $\lambda$ of the loss is backpropagated to $enc$. In our program we set $\lambda$ to 0.05. The loss for the ARAE can then be described by equation \ref{eqn:loss_arae}.
\begin{multline}\label{eqn:loss_arae}
\min\limits_{G} \max\limits_{D, enc} \mathcal{L}_{W}(D,G,enc) = \\ \E_{t\sim p_{d}}[ D(enc(t))] -  \E_{z\sim p_{z}}[ D(G(z))]  
\end{multline}
Both discriminator and generator are fully connected networks. The continuous representation is built with an autoencoder consisting of an LSTM for both encoder and decoder. The entire training process now consists of three steps: train the autoencoder on its reconstruction, train the discriminator and encoder to maximize the ability of the network to discriminate between real and generated samples, and finally, train the generator to try to fool the discriminator.
\begin{enumerate}
	\item $\min\limits_{enc,dec} \mathcal{L}_{rec}$
	\item  $\min\limits_{D,enc} \mathcal{L}_{D} = \max\limits_{D,enc}  \mathcal{L}_{W}(D,G,enc)$
	\item  $\min\limits_{G} \mathcal{L}_{G} = \min\limits_{G}  \mathcal{L}_{W}(D,G,enc)$
\end{enumerate}
The outcome of this setup is both the creation of continuous representations as well as the generation of realistic captions. The architecture is illustrated in figure \ref{fig:architecture}.

\subsection{Conditional ARAE} \label{captions}
\setlength{\belowcaptionskip}{-2pt}
\begin{table*}
	\centering
	\small
	\begin{tabular}{cc}
		\begin{tabular}{|p{2.2cm}|p{12.5cm}|}
			\hline
			\bf label & \bf generated caption \\
			\hline
			normal (+)&
			heart size within normal limits . no alveolar consolidation , no findings of pleural effusion or pulmonary edema . no pneumothorax . no pleural effusions .\\
			normal (-)&
			stable appearance of previous xxxx sternotomy . clear lungs bilaterally . redemonstration of disc disease of the thoracic spine . no pneumothorax or pleural effusion . clear lung volumes . \\
			cardiomegaly (+)&
			heart size is enlarged . stable tortuous aorta . no pneumothorax , pleural effusion or suspicious airspace opacity . prior granulomatous disease .  \\			
			cardiomegaly (-)&
			clear lungs . no infiltrates or suspicious pulmonary opacity . no pleural effusion or pneumothorax . cardiomediastinal silhouette within normal limits . calcified granulomas calcified granulomas . 		\\
			other (+)&		
			he heart size and pulmonary vascularity appear within normal limits . right pleural effusion is present and appears increased . the osseous structures appear intact . \\
			other (-)&		
			heart size and mediastinal contours are normal in appearance . no \textbf{oov} airspace consolidation . no pleural effusion or pneumothorax . the visualized bony structures are unremarkable in appearance . 
			\\\hline
		\end{tabular} & 
		
	\end{tabular}
	\caption{Examples of captions generated by the the conditional ARAE from a random vector $z$ and a class label. For each label an example of a correct (+) caption and a wrong (-) caption is given respectively.  }\label{tab:captions_cond}
\end{table*}

Additionally, the above setup of the ARAE is extended to allow content generation conditioned on input labels. From the MeSH labels in the dataset, we create three simple categories of diagnoses: normal, cardiomegaly and other (which includes a vast array of diagnoses ranging from pleural effusion, opacity, degenerative disease, and so on). 

During the training of the conditional ARAEs, the class or label is passed to both generator and discriminator. The formulation in equation ~\eqref{eqn:loss_arae} is supplemented by mentioning the conditional variable $c$ in $G(z,c)$ and $D(enc(t),c)$. During training the discriminator is presented with real samples $\hat{t}$ in combination with real labels $c_r$ as well as in combination with wrong labels $c_w$ that don't match the samples. Finally it is also presented with fake samples $\bar{t}$ in combination with the labels $c_r$. The discriminator is encouraged to learn that only the first combination is correct, while the generator tries to create samples that fool the discriminator given designated labels.

\section{Demonstration}

\begin{table}[t!]
	\begin{center}
		\begin{tabular}{|l|ccc|}
			\hline \bf  & \bf LSTM & \bf ARAE  & \bf ARAE   \\
			& & (uncondit.) & (condit.) \\
			
			\hline
			perplexity & 150.0 & 148.4 & 125.4\\		
			\hline
		\end{tabular}
	\end{center}
	\caption{\label{qual_eval} Perplexity scores for each of the models. }
\end{table}

In figure \ref{fig:demo_uncond} some captions are presented that are generated by the ARAE. Both during training and generation, sampling is performed with a temperature of 0.1. These examples qualitatively demonstrate that it is possible to generate text that mimics the complexity of medical reports.  

In table \ref{tab:captions_cond} we show some randomly chosen results for a network that produces text conditional on the class label. It becomes apparent that for the different labels, the network will produce wrong captions as well, especially for the label 'cardiomegaly' which has significantly less training examples. Empirically, the training is difficult and diverges a lot. We attribute the difficult convergence to two main factors. Firstly, despite the simple labels, the texts in different categories contain a large amount of overlap. Secondly, the conditional ARAE has many objective functions and four different networks ($enc, dec, G, D$) to optimize and balance in order to learn both what an informative representation looks like as well as how to generate it. 

In order to assess the performance of the system, we also train a baseline language model that consists of a 1-layer LSTM. The perplexity of the different models are presented in table \ref{qual_eval}. From the results we see that while both ARAE models achieve a lower perplexity than the LSTM, the conditional ARAE performs significantly better. 

We built a demo interface with the main goal of illustrating the quality and the diversity of the generated text\footnote{Demo available at https://liir.cs.kuleuven.be/software.php}. Upon starting the demo, the trained ARAE networks can be loaded by pressing 'Load\_models'. Once loaded, the user can interact in three ways. Firstly, sentences can be generated without conditional labels from a random noise vector by pressing 'Generate\_uncond'. Secondly, sentences can be generated from a random noise vector conditioned on one of the three labels (normal, cardiomegaly, other) by pressing 'Generate\_cond'. Finally, when pressing 'Discriminator\_game', a game starts where the user can attempt to fool the discriminator by inputting a short caption that might belong to an X-Ray. When the user presses 'enter', the system outputs whether the discriminator classifies it as a real caption or not. A screenshot of the interface is shown in figure \ref{fig:demo_uncond} where unconditional sentences were generated. 

\section{Conclusion}

With an Adversarially Regularized Autoencoder, a continuous text representation is learned of medical captions that can be useful in further applications. GANs are models that learn the underlying distribution while generating detailed continuous data. Therefore the successful training of a GAN on discrete data in the ARAE setup forebodes success for text generation as well. We illustrate the potential of GANs for discrete inputs by extending the ARAE architecture to create text conditioned on simple class labels, similar to conditional GANs. A quantitative evaluation shows that the conditional ARAE achieves a lower perplexity than both the unconditional ARAE and an LSTM baseline. 

\bibliography{naaclhlt2018}
\bibliographystyle{acl_natbib}

\appendix

\end{document}